\documentclass[sigconf]{acmart}
\AtBeginDocument{%
  \providecommand\BibTeX{{%
    \normalfont B\kern-0.5em{\scshape i\kern-0.25em b}\kern-0.8em\TeX}}}

\setcopyright{acmcopyright}
\copyrightyear{2022}
\acmYear{2022}
\acmDOI{XXXXXXX.XXXXXXX}

\acmConference[SETN 2022]{12th Hellenic Conference on Artificial Intelligence}{September 07-09, 2022}{Corfu, Greece}
\acmISBN{XXX-X-XXXX-XXXX-X/XX/XX}
\acmPrice{0.0}

\usepackage{textgreek}
\usepackage{microtype}
\usepackage{booktabs}
\usepackage{array}
\usepackage{multirow}
\usepackage{graphicx}
\usepackage{amsmath}
\usepackage{url}
\usepackage{comment}
\usepackage{xspace}
\usepackage{pifont}
\usepackage{enumitem}

\begin{document}

\title{Realistic Zero-Shot Cross-Lingual Transfer in Legal Topic Classification}


\author{Stratos Xenouleas [1,2], Alexia Tsoukara [2], Giannis Panagiotakis [2], \\ Ilias Chalkidis [1,2,3], Ion Androutsopoulos [1]}
\affiliation{%
\institution{[1] Athens University of Economics and Business, Greece}
\country{[2] Cognitiv+ Ltd, United Kingdom \quad [3] University of Copenhagen, Denmark}}


\renewcommand{\shortauthors}{Xenouleas et al.}

\newcommand{\eu}{\textsc{EU}\xspace}
\newcommand{\eurovoc}{\textsc{eurovoc}\xspace}
\newcommand{\eurlex}{\textsc{eurlex57k}\xspace}
\newcommand{\multieurlex}{\textsc{Multi-EURLEX}\xspace}

\newcommand{\eurlexen}{\textsc{eurlex57k-en}\xspace}
\newcommand{\eurlexde}{\textsc{eurlex57k-de}\xspace}
\newcommand{\eurlexfr}{\textsc{eurlex57k-fr}\xspace}
\newcommand{\eurlexel}{\textsc{eurlex57k-el}\xspace}

\newcommand{\multifit}{\textsc{MultiFiT}\xspace}
\newcommand{\opus}{\textsc{OPUS-MT}\xspace}
\newcommand{\mtfive}{\textsc{mT5}\xspace}
\newcommand{\mtwom}{\textsc{M2M}\xspace}
\newcommand{\laser}{\textsc{laser}\xspace}

\newcommand{\rprec}{\textit{R-PRC}\xspace}
\newcommand{\ndcg}{\textit{\textsc{ndcg}}\xspace}
\newcommand{\fone}{\textit{F1}\xspace}
\newcommand{\tr}{\textit{TR}\xspace}

\newcommand{\average}{\textsc{average}\xspace}
\newcommand{\lmtc}{\textsc{lmtc}\xspace}
\newcommand{\rpk}{\textit{RP}\xspace}
\newcommand{\rpfive}{\textit{RP@5}\xspace}
\newcommand{\pk}{\textit{P@k}\xspace}
\newcommand{\rk}{\textit{R@k}\xspace}
\newcommand{\dcg}{\textit{DCG}\xspace}
\newcommand{\ndcgk}{\textit{nDCG@k}\xspace}
\newcommand{\ndcgfive}{\textit{nDCG@5}\xspace}
\newcommand{\dcgk}{\textit{DCG@K}\xspace}
\newcommand{\idcgk}{\textit{IDCG@K}\xspace}

\newcommand{\xlmr}{\textsc{XLM-R}\xspace}
\newcommand{\xlm}{\textsc{xlm}\xspace}
\newcommand{\xlmroberta}{\textsc{XLM-R}\xspace}
\newcommand{\transformer}{\textsc{transformer}\xspace}
\newcommand{\transformers}{\textsc{transformers}\xspace}
\newcommand{\bert}{\textsc{BERT}\xspace}

\newcommand{\student}{\textit{Student}\xspace}
\newcommand{\teacher}{\textit{Teacher}\xspace}

\begin{abstract}
We consider zero-shot cross-lingual transfer in legal topic classification using the recent \multieurlex dataset. Since the original dataset contains parallel documents, which is unrealistic for zero-shot cross-lingual transfer, we develop a new version of the dataset without parallel documents. We use it to show that translation-based methods vastly outperform cross-lingual fine-tuning of multilingually pre-trained models, the best previous zero-shot transfer method for \multieurlex. We also develop a bilingual teacher-student zero-shot transfer approach, which exploits additional unlabeled documents of the target language and performs better than a model fine-tuned directly on labeled target language documents. 
\end{abstract}


\maketitle

\section{Introduction}
\label{sec:intro}
Transformer-based \citep{NIPS2017_3f5ee243} pre-trained language models \cite{devlin-etal-2019-bert} have significantly improved performance across NLP tasks. Multilingually pre-trained models~\cite{conneau-etal-2020-unsupervised, xue-etal-2021-mt5} have also been used for \emph{zero-shot cross-lingual transfer} \cite{hu2020xtreme, ruder-etal-2021-xtreme}, i.e., fine-tuning (further training) in one or more source languages and applying the model to other (unseen) target languages at inference. 
NLP for legal text has become popular~\cite{zhong-etal-2020-nlp,hendrycks2021cuad,chalkidis-etal-2021-multieurlex,lexglue,xiao2021}, but to our knowledge only \citet{chalkidis-etal-2021-multieurlex} have considered cross-lingual transfer of neural models in legal NLP. They introduced a multilingual dataset, \multieurlex, for legal topic classification and explored zero-shot cross-lingual transfer using multilingually pre-trained models like \xlmroberta \citep{conneau-etal-2020-unsupervised} combined with adaptation \citep{houlsby2019parameterefficient, zaken2021} to retain multilingual knowledge from pre-training. \multieurlex, however, contains to a large extent \emph{parallel} text (same content in multiple languages), which is unrealistic in real-world cross-lingual transfer. 
Also, \citet{chalkidis-etal-2021-multieurlex} did not consider \emph{translation-based} methods \cite{lample2019cross}, which machine-translate the target language documents to a source language, or machine-translate the labeled source documents to the target languages and use the translations to train models for the target languages. \emph{Teacher-student} approaches, which leverage multilingual teacher models to soft-label unlabeled documents of the target language(s) to train a student  \cite{eisenschlos-etal-2019-multifit}, were also not considered. We address these limitations in this work.

\vspace{-1mm}
\begin{itemize}[leftmargin=8pt]
\setlength\itemsep{0em}
\item We construct, use, and release a new, \textbf{more realistic version of \multieurlex} that contains non-parallel training documents in four languages (English, French, German, Greek), along with the same (parallel) development and test documents for those languages as in the original dataset.
\item To establish \textbf{`upper' performance bounds} for zero-shot transfer, we fine-tune \xlmroberta separately per language, as well as jointly in all four languages, simulating a scenario where there are equally many training documents in all languages, also confirming that adapters improve cross-lingual transfer. Unlike \citet{chalkidis-etal-2021-multieurlex}, we find that jointly fine-tuning for all languages leads to better performance, compared to monolingual fine-tuning. We partly attribute this difference to the fact that the original dataset contains parallel documents (same content), which reduces the benefit of jointly training in multiple languages.
\item We show that \textbf{translation-based methods \emph{vastly} outperform cross-lingual fine-tuning} with adapters, which was the best zero-shot cross-lingual transfer method of \citet{chalkidis-etal-2021-multieurlex}. This suggests that exploiting modern Neural Machine Translation (NMT) systems is a much better zero-shot cross-lingual transfer strategy in real life, at least for legal topic classification.
\item We develop a \textbf{bilingual teacher-student}. A multilingually pre-trained teacher is fine-tuned on labeled documents of the source language and their machine-translations in the target language. The teacher then soft-labels all the documents it was trained on, and also soft-labels unlabeled documents of the target language. A student is then trained to predict all the soft labels. Its performance \textbf{exceeds the monolingual `upper bound', i.e., fine-tuning directly in the target language}. Also, the student supports both the target and the source language, which allows a company to support \textbf{both languages with a single model}.
\end{itemize}

\section{Related Work}
Pre-trained Transformers have boosted performance across NLP, including cross-lingual transfer \citep{conneau-2019-xlm,conneau-etal-2020-unsupervised, xue-etal-2021-mt5}. Adapter modules \citep{houlsby2019parameterefficient} have been used to transfer pre-trained models to low-resource or even unseen languages \citep{pfeiffer-etal-2020-mad, pfeiffer-etal-2021-unks}. Also, \citet{eisenschlos-etal-2019-multifit} proposed \multifit, a teacher-student framework that allows pre-training and fine-tuning monolingual students  in a target language, using a multilingually pre-trained teacher to bootstrap the student with soft-labeled  documents of the target language.

\citet{gonalves2010} performed legal topic classification in English, German, Italian,  Portuguese using monolingual SVMs and their combination as a multilingual ensemble. \citet{chalkidis-etal-2021-multieurlex} studied zero-shot cross-lingual transfer in legal topic classification, introducing \multieurlex. They found that fine-tuning a multilingually pretrained model in a single language leads to catastrophic forgetting of the multilingual knowledge from the pre-training and, thus, performs poorly in zero-shot transfer to other languages. To retain the multilingual knowledge, they used adaptation strategies \citep{houlsby2019parameterefficient, pfeiffer-etal-2020-mad}. Their results also show that zero-shot cross-lingual transfer is more challenging in legal topic classification, compared to more generic classification tasks~\cite{hu2020xtreme, ruder-etal-2021-xtreme}.

\section{The New \multieurlex Version}
We use \multieurlex \citep{chalkidis-etal-2021-multieurlex}, a multilingual dataset for legal topic classification comprising 65k EU laws officially translated in 23 \eu languages.\footnote{\multieurlex is available at \url{https://huggingface.co/datasets/multi_eurlex}. Our modified version is available at \url{https://huggingface.co/datasets/nlpaueb/multi_eurlex}.} Each document (\eu law) was originally annotated with relevant EUROVOC\footnote{\url{http://eurovoc.europa.eu/}} concepts by the Publications Office of \eu. EUROVOC is a taxonomy of concepts (a hierarchy of labels). We use the 127 `Level 2' labels, obtained by \citet{chalkidis-etal-2021-multieurlex} from the original EUROVOC annotations of the documents.\vspace{1mm}

\noindent\textbf{Limitations of Multi-EURLEX:}
One limitation of Multi-EURLEX is that the number of training documents is not the same across languages. For languages spoken in the older \eu member states, there are 55k training documents per language, but for many others,
there are much fewer training documents (e.g., 8k for Croatian, 15k for Bulgarian). This makes zero-shot cross-lingual transfer results difficult to compare, because the training set size varies across experiments, a factor not controlled for by \citet{chalkidis-etal-2021-multieurlex}. More importantly, when training in several source languages, most of the source language documents are parallel (same content in multiple languages), which is unrealistic in most real-life applications and may produce misleading results. For example, in one of their baselines, \citet{chalkidis-etal-2021-multieurlex} jointly fine-tune a multilingually pre-trained model on the (parallel) training documents of all the 23 languages, and observe no performance benefit compared to fine-tuning a different instance of the model per language, possibly because of the fact that the training documents are parallel (same content). By contrast, we find that the multilingually fine-tuned model is substantially better than the monolingual ones, when the training documents are not parallel.\vspace{1mm}

\noindent\textbf{Updated Harder Version:}
We, therefore, construct, use, and release a new, more realistic version of \multieurlex, where there are no parallel training documents across languages. For the new version, we randomly selected 12k (11k training, 1k development) documents per language, limiting the languages to four, namely English, German, French, Greek, and making sure there are no parallel documents. Using four languages allowed us to avoid parallel documents, but still have a reasonably large training set (11k) per language. The test sets are still parallel (5k training per language, as in the original \multieurlex) to allow comparisons to be made when changing the target language. The four languages are from three different families (Germanic, Romance, Hellenic), which makes cross-lingual transfer harder.

\begin{table*}[ht!]
    \centering
    \resizebox{0.9\textwidth}{!}{
        \begin{tabular}{llcc|c|ccc|c}
            \hline
            & & &  & \bf Source & \multicolumn{3}{c|}{\bf Target Languages} & \bf Target\\
             \bf Model & \bf \#M & \bf MT & \bf BS+SL  & \bf en & \bf de & \bf fr & \bf el & \bf Avg\\
             \hline
             \hline
            \multicolumn{9}{c}{\textit{\textbf{`Upper' performance bounds (labeled training documents available in all 4 languages)}}} \\
            \hline
            \hline
             \multicolumn{9}{l}{\textit{\textbf{Monolingual FT (Fine-Tuning on labeled  documents of a particular language only)}}} \\ 
             \hline
             \xlmr (E2E)    & 4 & \ding{55} & \ding{55} & 68.2 \footnotesize{$\pm$ 0.8}  &  65.8 \footnotesize{$\pm$ 0.7}  &  67.0 \footnotesize{$\pm$ 1.7}  &  64.6 \footnotesize{$\pm$ 0.4}  &  65.8 \\
             \xlmr +Adapters   & 4 & \ding{55}  & \ding{55} & 68.8 \footnotesize{$\pm$ 0.1}  &  65.0 \footnotesize{$\pm$ 0.7}  &  68.1 \footnotesize{$\pm$ 0.4}  &  64.9 \footnotesize{$\pm$ 0.2}  &  66.0 \\ 
             \hline
             \multicolumn{9}{l}{\textit{\textbf{Multilingual FT (jointly Fine-Tuning on labeled documents of all 4 languages)}}} \\ 
             \hline
             \xlmr (E2E)       & 1  & \ding{55} & \ding{55} & 70.0 \footnotesize{$\pm$ 1.0}  &  68.9 \footnotesize{$\pm$ 1.0}  &  69.1 \footnotesize{$\pm$ 1.5}  &  67.4 \footnotesize{$\pm$ 0.6}  &  68.5 \\
             \xlmr +Adapters     & 1  & \ding{55} & \ding{55} & 70.4 \footnotesize{$\pm$ 1.6}  &  69.2 \footnotesize{$\pm$ 1.1}  &  69.9 \footnotesize{$\pm$ 1.6}  &  67.1 \footnotesize{$\pm$ 0.5}  &  68.7 \\
             \hline
             \hline
             \multicolumn{9}{c}{\textit{\textbf{Zero-shot Cross-lingual Methods (no labeled training documents available in the Target languages)}}} \\
             \hline
             \hline
             \multicolumn{9}{l}{\textit{\textbf{Cross-lingual FT (FT on Source documents only, test in each Target language directly)}}} \\ 
             \hline
             \xlmr (E2E)        & 1  & \ding{55} & \ding{55} & ---               &  55.2 \footnotesize{$\pm$ 5.2}  &  58.1 \footnotesize{$\pm$ 2.9}  &  42.8 \footnotesize{$\pm$ 6.5}  &  52.0 \\ 
             \xlmr +Adapters       & 1  & \ding{55} & \ding{55} & ---               &  61.7 \footnotesize{$\pm$ 1.9}  &  60.6 \footnotesize{$\pm$ 0.8}  &  48.1 \footnotesize{$\pm$ 1.8}  &  56.8 \\ 
             \hline
             \multicolumn{9}{l}{\textit{\textbf{Translate Test (FT on Source documents only, test on Target documents translated to Source)}}} \\
             \hline
             \xlmr (E2E)      & 1  & \ding{51} & \ding{55} & ---	           &  63.3 \footnotesize{$\pm$ 1.8}  &  68.1 \footnotesize{$\pm$ 0.8}  &  66.5 \footnotesize{$\pm$ 1.0}  &  66.0 \\ 
             \xlmr +Adapters      & 1  & \ding{51} & \ding{55} & ---	           &  62.8 \footnotesize{$\pm$ 1.0}  &  \textbf{68.7} \footnotesize{$\pm$ 0.2}  &  67.2 \footnotesize{$\pm$ 1.2}  &  66.2 \\ 
             \hline
             \multicolumn{9}{l}{\textit{\textbf{Translate Train (translate the Source training documents to each Target, FT on the translations)}}} \\ 
             \hline       
             \xlmr (E2E)   & 4  & \ding{51}  & \ding{55} & ---	             &  66.7 \footnotesize{$\pm$ 1.5}  &  67.2 \footnotesize{$\pm$ 1.1}  &  64.1 \footnotesize{$\pm$ 1.4}  &  66.0 \\
             \xlmr +Adapters  & 4  & \ding{51}  & \ding{55} & ---	             &  67.2 \footnotesize{$\pm$ 1.0}  &  67.0 \footnotesize{$\pm$ 1.2}  &  64.8 \footnotesize{$\pm$ 1.7}  &  66.4 \\
             \hline 
             \multicolumn{9}{l}{\textit{\textbf{Monolingual Teacher-Student (jointly FT on translations from Src to Target language and extra docs in Target)}}} \\ 
             \hline 
             \xlmr (E2E) (Student)   & 4  & \ding{51} & \ding{51} & ---  & 65.9 \footnotesize{$\pm$0.4}	& 68.0 \footnotesize{$\pm$1.1}  &	48.6 \footnotesize{$\pm$0.6} & 60.7 \\
             \hline 
             \multicolumn{9}{l}{\textit{\textbf{Bilingual Teacher-Student (jointly FT on Source documents and their translations in a Target language)}}} \\ 
             \hline 
             \xlmr (E2E) (Student)   & 4  & \ding{51} & \ding{51} & \textbf{69.1} \footnotesize{$\pm$ 1.3}  & \textbf{67.4} \footnotesize{$\pm$ 0.1}  &  66.1 \footnotesize{$\pm$ 0.3}  &  65.0 \footnotesize{$\pm$ 0.4}  &  66.1 \\
             \hline
             \xlmr +Adapters (Student)  & 4  & \ding{51} & \ding{51} & 67.8 \footnotesize{$\pm$ 1.3}  &  66.9 \footnotesize{$\pm$ 0.3}  &  67.6 \footnotesize{$\pm$ 1.2}  &  \textbf{67.9} \footnotesize{$\pm$ 0.1}  &  \bf 67.5 \\
             \hline
             \multicolumn{9}{l}{\textit{\textbf{Multilingual Teacher-Student (jointly FT on Source documents and their translations in all Target languages)}}} \\ 
             \hline
             \xlmr (E2E) (Student)    & 1  & \ding{51} & \ding{51} & 62.3 \footnotesize{$\pm$ 1.6}  &  60.9 \footnotesize{$\pm$ 0.3}  &  66.8 \footnotesize{$\pm$ 0.2}  &  48.4 \footnotesize{$\pm$ 0.3}  &  58.7 \\
             \hline
             \xlmr +Adapters (Student)  & 1  & \ding{51}  & \ding{51} & 65.0 \footnotesize{$\pm$ 0.2}  &  62.6 \footnotesize{$\pm$ 0.2}  &  \textbf{68.7} \footnotesize{$\pm$ 0.8}  &  50.5 \footnotesize{$\pm$ 0.0}  &  60.6 \\
             \bottomrule
        \end{tabular}
    }
    \caption{Test R-Precision ($\mathrm{RP}$, \%) results $\pm$ std.\ deviation over 3 runs with different random seeds. E2E: End-to-End Fine-Tuning (FT). +Adapters: Updating only Adapter layers and classification head during FT. \#M: number of models fine-tuned. MT: machine-translated documents used. BS+SL: Boot-Strapping with Soft Labels.
    }
    \label{tab:results}
    \vspace{-8mm}
\end{table*}

\section{Experimental Setup and Methods}
\label{sec:methods}

We experiment with \xlmr \cite{conneau-etal-2020-unsupervised} in the two best-performing configurations of \citet{chalkidis-etal-2021-multieurlex}: (a) \emph{End-to-end} (E2E) fine-tuning, where all model parameters are updated, and (b) \emph{Adapter-based}~\cite{houlsby2019parameterefficient} fine-tuning, where we only update the parameters of additional bottleneck (adapter) layers between the pre-trained Transformer blocks. We compare both configurations across several settings:\vspace{2mm}

\noindent\textbf{`Upper' Performance Bounds:}
Firstly, we examine the performance of \xlmroberta fine-tuned in a \emph{monolingual} fashion, i.e., separately on the labeled documents of each language (source or target), or in a \emph{multilingual} fashion, i.e.,  jointly on training documents of all four languages. In real life, labeled data in the target languages are rarely available. Typically a company has trained a system on English labeled documents and wishes to deploy it in other languages with very few (or no) labeled documents. However, these experiments show how high performance would be in an ideal case with labeled documents in each target language (as many as in the source language). We call them an `upper' bound, because we would expect performance to be inferior in zero-shot cross-lingual transfer, where no labeled documents are available in the target languages. Nevertheless, our best zero-transfer method, actually surpasses some `upper' bounds.\vspace{2mm}

\noindent\textbf{Cross-lingual Fine-Tuning (FT):}
\citet{chalkidis-etal-2021-multieurlex} showed that when fine-tuning a multilingually pre-trained model for a particular language, the model largely `forgets' its knowledge of the other languages and performs poorly in zero-shot cross-lingual transfer, unless adaptation mechanisms are used; but even in the latter case, zero-shot performance was much lower than the `upper' bounds.\vspace{2mm}

\noindent\textbf{Translation-based Methods:}
Following \citet{conneau-etal-2020-unsupervised} and \citet{xue-etal-2021-mt5}, we also consider methods that exploit machine-translated documents.\footnote{We use the EasyNMT~\cite{easynmt_2021} framework.
} In \textit{Translate Test}, we fine-tune \xlmr for the source language; given a target language document at inference time, we simply translate it to the source language and use the fine-tuned (for the source language) \xlmr. In \textit{Translate Train}, we machine-translate the labeled training documents of the source language to the target language, and use the translations (and the original labels) to fine-tune \xlmr for the target language; at test time, we evaluate on labeled test documents written in the target language (not machine-translated).\vspace{2mm}

\noindent\textbf{Teacher-Student:} Inspired by \citet{eisenschlos-etal-2019-multifit}, we first fine-tune a \emph{bilingual teacher} \xlmr using labeled documents in the source language and their machine translations (and original labels) in the target language. Then, we use the teacher to soft-label (assign a probability per label to) the source and machine-translated documents it was trained on, and to soft-label additional unlabeled documents of the target language; we use the 12k training documents of the target language without their labels. 
We then train a \emph{student} \xlmr (on all the documents the teacher soft-labeled) to predict the soft labels. The student (and the teacher) is bilingual, i.e., it supports both the target and the source language. This allows a company to support both languages with a single model, which has cost benefits. 
We also experiment with a \emph{multilingual teacher-student} approach, where a single multi-lingual teacher is jointly fine-tuned on labeled documents of the source language and their machine translations in all target languages. The  teacher then soft-labels all the documents (and translations) it was trained on and additional unlabeled documents of the target languages. The student is again trained to predict the soft labels.\footnote{The student sees soft labels even in the manually labeled target documents and their translations, since soft labels have been found beneficial in manually labeled documents too \cite{fornaciari-etal-2021-beyond}. Preliminary experiments confirmed this.} In this case, all four languages are supported.\vspace{-1mm}

\section{Experiments} 
\label{sec:results}
Table~\ref{tab:results} reports test results when the source language is English. The same conclusions can be drawn with other source languages (French, German, Greek); see Appendix~\ref{sec:extra_results}.  Following \citet{chalkidis-etal-2021-multieurlex}, we report average R-Precision ($\mathrm{RP}$) \cite{manning2009} alongside ($\pm$) standard deviation over 3 runs with different random seeds on the test set. We report results both per language (English, German, French, Greek) and on average across all languages. The updated code base is available on Github.\footnote{\url{https://github.com/nlpaueb/multi-eurlex/tree/realistic-zero-shot}}\vspace{-2mm}

\subsection{Main Experimental Results \& Analysis}
\label{sec:main_experiments}

Starting from the `upper' bound results, we find that jointly fine-tuning on all four languages performs substantially better than fine-tuning monolingual models. By contrast, \citet{chalkidis-etal-2021-multieurlex} reported no benefit when jointly fine-tuning \xlmr for multiple languages. However, in their experiments there were many more training documents per language and the documents were parallel translations (same content), which reduced the benefit of jointly training in multiple languages (in our case, four times more documents with different content). 
Cross-lingual FT with Adapters performs approx.\ 10 points lower in the target languages on average, compared to the corresponding monolingual `upper' bound (56.8 vs.\ 66.0).
Translate Test and Train, which were not considered by \citet{chalkidis-etal-2021-multieurlex}, vastly outperform Cross-lingual FT with Adapters, which was the best zero-shot method of \citet{chalkidis-etal-2021-multieurlex}, and perform on par with the monolingual `upper' bounds.

We identify three main factors that potentially affect the classification performance per target language: (a) the \emph{general capability (`trends') of \xlmr} in a given language, which can be estimated given generic benchmarking of \xlmr \cite{conneau-etal-2020-unsupervised} and the `upper' bound results; (b) the \emph{quality of machine translations}, when machine-translations are used; and (c) \emph{temporal concept drift}, which can be estimated by the label distribution alignment between the train and test subsets. We present a preliminary analysis of these factors and how they could affect  classification performance, leaving for future work a further study of the factors and how they interact.\vspace{1mm}

\noindent\textbf{General capability (`trends') of \xlmr:}
Based on the findings of \citet{conneau-etal-2020-unsupervised} and \citet{chalkidis-etal-2021-multieurlex}, we observe that the classification performance of \xlmr per language correlates with our own empirical results, with English being the top-performing language, followed by French, German, and Greek in order. This is highly expected since \xlmr is not equally pre-trained across all languages in terms of training examples (despite the use of exponential smoothing in data sampling), and vocabulary coverage, i.e., more Latin-based sub-words compared to ones in Greek script.

\begin{table}[ht!]
    \centering
    \resizebox{0.9\columnwidth}{!}{
    \begin{tabular}{cc|cc|cc}
    \toprule
        \multicolumn{2}{c}{\bf en-to-de} & \multicolumn{2}{c}{\bf en-to-fr} & \multicolumn{2}{c}{\bf en-to-el}\\
        train & test & train & test & train & test \\
        0.680 & 0.728 & 0.733 & 0.803 & 0.680 & 0.720 \\
        \bottomrule
    \end{tabular}
    }
    \caption{Quality of machine translations, from source English (en) to target languages (German--de, French--fr, Greek--el), measured in terms of METEOR scores.}
    \label{tab:nmt}
    \vspace{-6mm}
\end{table}

\noindent\textbf{Translation Quality:}
Table~\ref{tab:nmt} reports the quality of machine translations measured in terms of METEOR \cite{banerjee-lavie-2005-meteor}, using as references the original human translations. We observe that the quality from English to French (0.73) is substantially better than from English to German or Greek (0.68), when considering the training documents; similar results are obtained from the (parallel) test documents. This quality disparity could potentially affect the performance of all methods that use machine translated documents, i.e., translate-train, translate-test, bilingual/multilingual teacher-student. Indeed, we observe in Table~\ref{tab:results} that these methods are consistently better in French, comparable in German, and worse in Greek. This is quite expected as both French and German use the Latin alphabet, and share a larger part of the subword vocabulary, compared to Greek.\vspace{1mm}

\begin{figure}[h]
\centering
\includegraphics[width=\columnwidth]{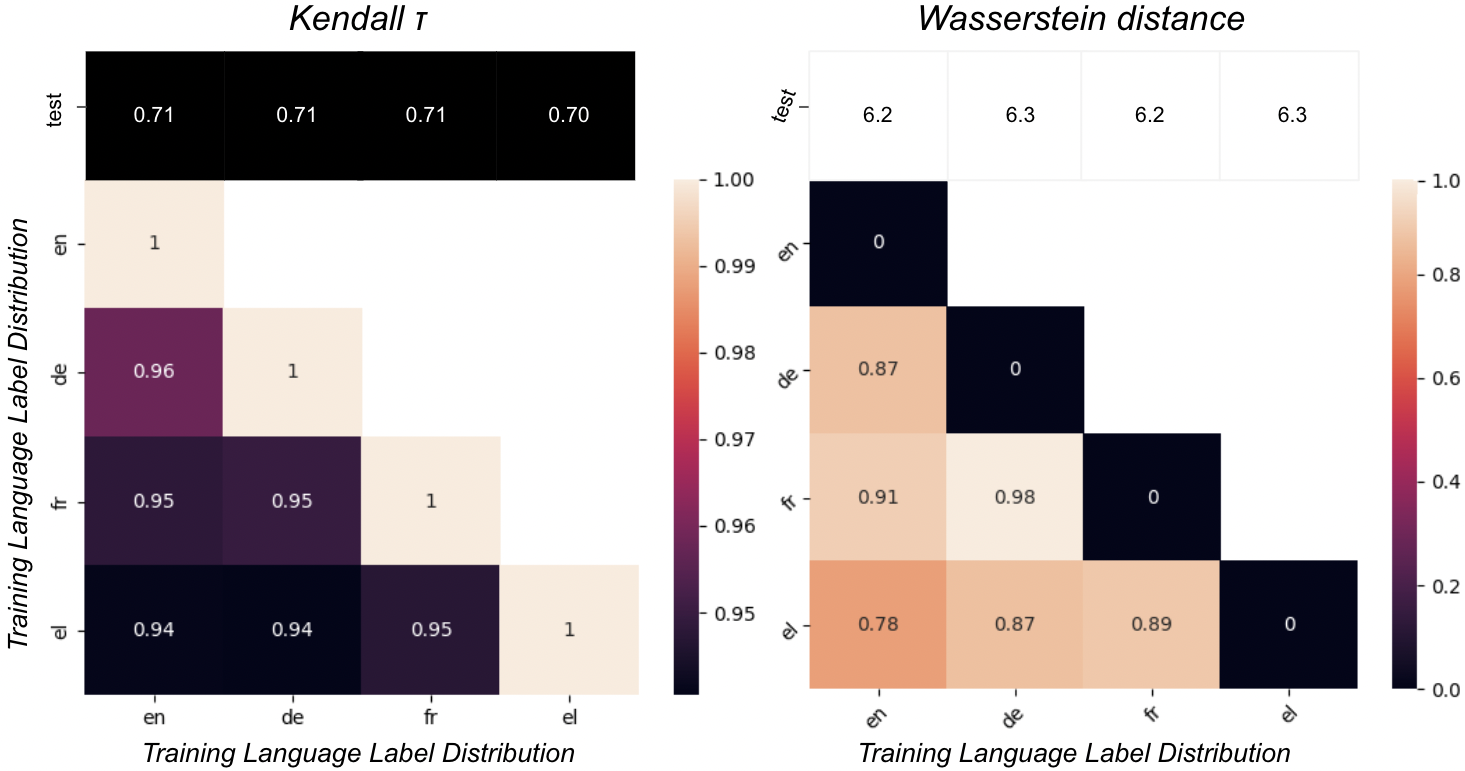}
\vspace{-6mm}
\caption{Kendall $\tau$ ($\uparrow$ , higher is better) and Wasserstein distance ($\downarrow$ , lower is better) between training (or test) label distributions per language pair.}
\label{fig:measures}
\vspace{-4mm}
\end{figure}

\noindent\textbf{Temporal Concept Drift:} To avoid parallel data, in the new dataset the training sets of different languages come comprise different documents. This introduces a temporal concept drift with respect to the time period of the shared (parallel) test set. 
To study this drift, we use two alternative measures, Kendall's $\tau$ and Wasserstein distance. In both cases, we measure the distance between the training label distributions of two different languages (e.g., en-el), or the distance between the shared test label distribution and the training label distribution of a language (e.g., test-en). Kendall's $\tau$ is only sensitive to the label rankings of the compared distributions (labels ranked by frequency), while Wasserstein distance is more sensitive, in that it also considers the frequency differences of the labels across the two distributions. In Figure~\ref{fig:measures}, we observe that both measures favor the English-to-French (en-fr) and English-to-German (en-de) language pairs, compared to English-to-Greek (en-el); the difference is larger when measured by the more sensitive Wasserstein distance. The difference between the Greek training label distribution and the test distribution is also larger compared to the other languages.

\subsection{Teacher-Student Experimental Results}

The monolingual student is comparable with the monolingual `upper' bound with adapters in German and French, but it has lower performance in Greek. This performance disparity may be the result of the general factors described in Section~\ref{sec:main_experiments}, which disfavor Greek compared to the rest of the languages.

The bilingual student with Adapters improves the average performance on target languages slightly further (67.5), exceeding the monolingual `upper' bound with Adapters (66.0). This improvement can be attributed to the additional (originally unlabeled) documents of the target languages and the soft labels that the student uses. Recall that the student model has the further practical advantage of supporting two languages. 

The multilingual student performs much worse on average, compared to the bilingual student, even with Adapters; with the exception of French, where the student performs best (68.7) compared to all other methods. These results seem to be related to the general factors described in Section~\ref{sec:main_experiments} and the quality of the teacher's soft labels, but further work is needed to study them.
As a first step, in Appendix~\ref{sec:main_analysis}
we provide an initial analysis of the quality of the soft labels per language and document subset (source or target language, human- or machine-translated, originally labeled or unlabeled), which seems to be in line with the performance of the multilingual student in Table~\ref{tab:results}.

\section{Conclusions and Future Work}
We considered zero-shot cross-lingual transfer in legal topic classification, introducing a more realistic version of \multieurlex without parallel documents. We showed that translation-based methods vastly outperform cross-lingual fine-tuning of multilingually pre-trained models, the best previous zero-shot transfer method for \multieurlex. We also developed a bilingual teacher-student zero-shot transfer approach, which exploits additional unlabeled documents of the target language and performs better than a model fine-tuned directly on labeled target language documents, while supporting both languages with a single model.

In future work, we aim to better understand the reasons of the poor performance of the \emph{multilingual} teacher-student and hopefully to address them, in order to deploy a single zero-shot cross-lingual transfer model for multiple target languages.

\section*{Acknowledgments}

This work is partly funded by the Innovation Fund Denmark (IFD)\footnote{\url{https://innovationsfonden.dk/en}} under File No.\ 0175-00011A. This research has been also co‐financed by the European Regional Development Fund of the European Union and Greek national funds through the Operational Program Competitiveness, Entrepreneurship and Innovation, under the call RESEARCH – CREATE – INNOVATE (Τ2ΕΔΚ-03849).

\bibliographystyle{ACM-Reference-Format}
\bibliography{anthology,custom}

\appendix

\begin{figure*}[ht]
    \centering
    \includegraphics[width=0.9\textwidth]{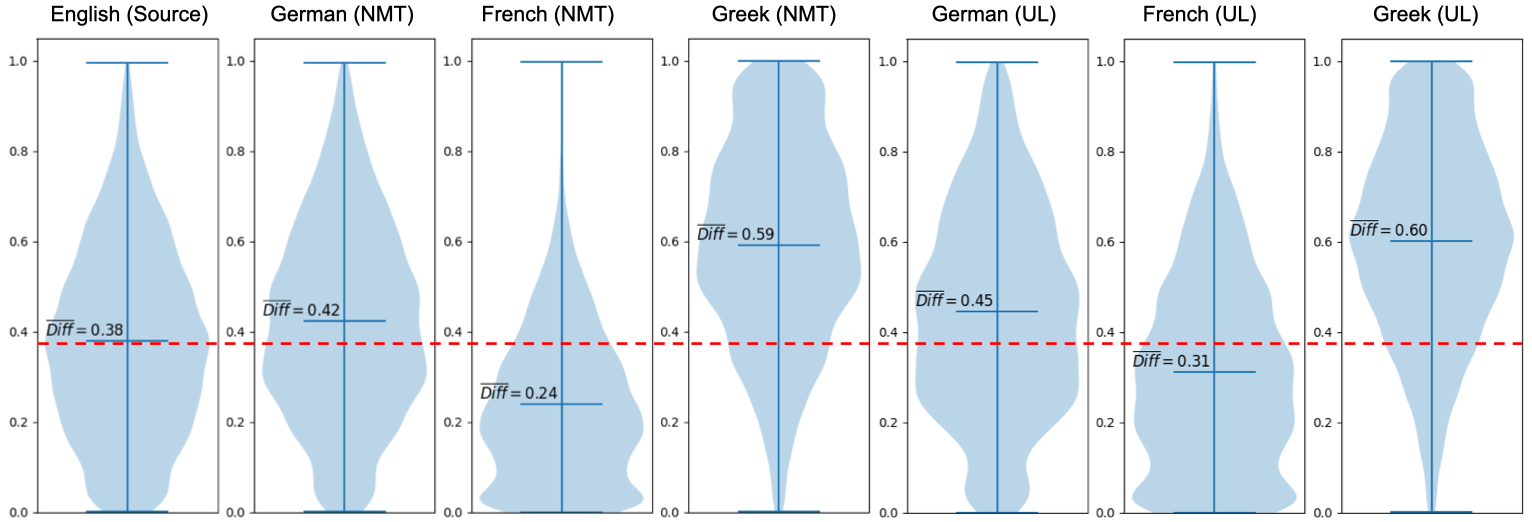}
    \vspace{-2mm}
    \caption{Average difference ($\overline{\mathrm{Diff}}$)
    between \emph{gold} and \emph{soft} labels (the latter predicted by the multilingual teacher), measured as Mean Absolute Error averaged over the document subset being considered. Results reported per document subset: original in English (source), machine-translated (NMT) in target languages, and additional unlabelled (UL) in the target languages.}
    \label{fig:main_violins}
    \vspace{-2mm}
\end{figure*}

\section{Additional Details}

\noindent\textbf{Experimental Details} We use the best hyper-parameters reported by \citet{chalkidis-etal-2021-multieurlex}.  For end-to-end (E2E) fine-tuning with XLM-R, we use a learning rate of 3e-5 and batch size 8. With adapter modules, we use a learning rate of 1e-4 and bottleneck size 256. For additional details consider Appendix A of \citet{chalkidis-etal-2021-multieurlex}.\vspace{2mm}

\noindent\textbf{Computational Details}
All experiments ran on an \textsc{NVIDIA DGX-1} station with 8 \textsc{NVIDIA V}100 16\textsc{GB GPU} cards; each experiment run in a single \textsc{GPU} card. Table~\ref{tab:model_runtime} shows the run-time (training until convergence) of every experiment across 3 runs with different random seeds.\vspace{2mm}

\begin{table}[h]
    \centering
    \resizebox{\columnwidth}{!}{
        \begin{tabular}{l|c|c}
        \toprule
            \bf Setting & \bf Adapters & \bf Avg Run Time \\
            \midrule
           Monolingual & \ding{55} & 2h \\
           Monolingual & \ding{51} & 4h \\
           \midrule
           Multilingual & \ding{55} & 5h \\
           Multilingual & \ding{51} & 9h \\
           \midrule
           Cross-lingual + MT & \ding{55} & 2h \\
           Cross-lingual + MT & \ding{51} & 4h \\
           \midrule
           Bilingual (Teacher)  & \ding{55} & 13h \\
           Bilingual (Student) & \ding{51} & 10h \\
           \midrule
           Multilingual (Teacher) & \ding{55} & 18h \\
           Multilingual (Student) & \ding{51} & 15h \\
           \bottomrule
         \end{tabular}
    }
    \caption{Run-time (training until convergence) of every experiment (each experiment running on a single Tesla V100 GPU) averaged over 3 runs with different random seeds.}
    \label{tab:model_runtime}
    \vspace{-8mm}
\end{table}

\noindent\textbf{Translation Details}
For machine translation, we used the EasyNMT\footnote{\url{https://github.com/UKPLab/EasyNMT}} framework utilizing the \emph{many-to-many} M2M\_100\_418M model of \citet{fan2020} for el-to-en and el-to-de pairs and the OPUS-MT \cite{tiedemann-thottingal-2020-opus} models for the rest. A manual check of some translated samples showed sufficient translation quality.

\begin{table*}[ht!]
    \centering
    \resizebox{0.8\textwidth}{!}{
        \begin{tabular}{llcc|c|ccc|c}
        \hline
            & & &  & \bf Source & \multicolumn{3}{c|}{\bf Target Languages} & \bf Target\\
             \bf Model & \bf \#M & \bf MT & \bf BS+SL  & \bf de & \bf en & \bf fr & \bf el & \bf Avg\\
             \hline
             \hline
             \multicolumn{9}{c}{\textit{\textbf{Zero-shot Cross-lingual FT (No labeled data in target languages)}}} \\
             \hline
             \hline
             \multicolumn{9}{l}{\textit{\textbf{Cross-lingual FT (German Only)}}} \\ 
             \hline
             \xlmr              & 1  & \ding{55} & \ding{55} &  65.84 \footnotesize{$\pm$ 0.68}  & 57.43 \footnotesize{$\pm$ 1.61}  &  53.95 \footnotesize{$\pm$ 2.48}  &  44.97 \footnotesize{$\pm$ 1.09}  &  52.1 \\ 
             \xlmr + Adapters   & 1  & \ding{55} & \ding{55} &  64.98 \footnotesize{$\pm$ 0.72}  & 61.30 \footnotesize{$\pm$ 1.70}  &  58.28 \footnotesize{$\pm$ 0.60}  &  49.02 \footnotesize{$\pm$ 1.09}  &  56.2 \\ 
             \hline
             \multicolumn{9}{l}{\textit{\textbf{Translate Test documents to Target language}}} \\
             \hline
             \xlmr              & 1  & \ding{51} & \ding{55} &  65.84 \footnotesize{$\pm$ 0.68}  & 65.65 \footnotesize{$\pm$ 0.72}  &  65.66 \footnotesize{$\pm$ 0.78}  &  63.57 \footnotesize{$\pm$ 0.74}  &  65.0 \\ 
             \xlmr + Adapters   & 1  & \ding{51} & \ding{55} &  64.98 \footnotesize{$\pm$ 0.72}  & 65.66 \footnotesize{$\pm$ 1.16}  &  64.76 \footnotesize{$\pm$ 0.50}  &  64.70 \footnotesize{$\pm$ 1.61}  &  65.0 \\ 
             \hline
             \multicolumn{9}{l}{\textit{\textbf{Translate Train documents to Target language}}} \\ 
             \hline       
             \xlmr              & N  & \ding{51}  & \ding{55} &  65.84 \footnotesize{$\pm$ 0.68}  & 67.36 \footnotesize{$\pm$ 1.62}  &  65.64 \footnotesize{$\pm$ 1.14}  &  64.32 \footnotesize{$\pm$ 1.21}  &  65.8 \\
             \xlmr + Adapters   & N  & \ding{51}  & \ding{55} &  64.98 \footnotesize{$\pm$ 0.72}  & 66.03 \footnotesize{$\pm$ 1.40}  &  65.74 \footnotesize{$\pm$ 1.53}  &  63.85 \footnotesize{$\pm$ 0.18}  &  65.2 \\
             \bottomrule
        \end{tabular}
    }
    \caption{Test R-Precision ($\mathrm{RP}$, \%) results $\pm$ std.\ deviation over 3 runs with different random seeds. E2E: End-to-End Fine-Tuning (FT). +Adapters: Updating only Adapter layers and classification head during FT. \#M: number of models fine-tuned. MT shows if machine-translated documents are used. BS+SL shows if teacher-student Boot-Strapping with Soft Labels is used.}
    \label{tab:appendix_de_results}
\end{table*}

\begin{table*}[ht!]
    \centering
    \resizebox{0.8\textwidth}{!}{
        \begin{tabular}{llcc|c|ccc|c}
        \hline
             & & & & \bf Source & \multicolumn{3}{c|}{\bf Target Languages} & \bf Target\\
             \bf Model & \bf \#M & \bf MT & \bf BS+SL  & \bf fr & \bf en & \bf de & \bf el & \bf Avg\\
             \hline
             \hline
             \multicolumn{9}{c}{\textit{\textbf{Zero-shot Cross-lingual FT (No labeled data in target languages)}}} \\
             \hline
             \hline
             \multicolumn{9}{l}{\textit{\textbf{Cross-lingual FT (French Only)}}} \\ 
             \hline
             \xlmr              & 1  & \ding{55} & \ding{55} &  67.01 \footnotesize{$\pm$ 1.69}  & 65.26 \footnotesize{$\pm$ 0.85}  &  57.04 \footnotesize{$\pm$ 2.74}   &  49.27 \footnotesize{$\pm$ 2.17}  &  57.2 \\ 
             \xlmr + Adapters   & 1  & \ding{55} & \ding{55} &  68.05 \footnotesize{$\pm$ 0.35}  & 64.98 \footnotesize{$\pm$ 1.66}  &  61.44 \footnotesize{$\pm$ 1.80}   &  51.31 \footnotesize{$\pm$ 1.86}  &  59.2 \\ 
             \hline
             \multicolumn{9}{l}{\textit{\textbf{Translate Test documents to Target language}}} \\
             \hline
             \xlmr              & 1  & \ding{51} & \ding{55}  &  67.01 \footnotesize{$\pm$ 1.69}  & 66.73 \footnotesize{$\pm$ 1.86}	&  59.49 \footnotesize{$\pm$ 2.26}   &  46.16 \footnotesize{$\pm$ 0.42}  &  57.5 \\ 
             \xlmr + Adapters   & 1  & \ding{51} & \ding{55}  &  68.05 \footnotesize{$\pm$ 0.35}  & 66.72 \footnotesize{$\pm$ 1.11}	&  59.59 \footnotesize{$\pm$ 0.24}   &  46.98 \footnotesize{$\pm$ 2.56}  &  57.8 \\ 
             \hline
             \multicolumn{9}{l}{\textit{\textbf{Translate Train documents to Target language}}} \\ 
             \hline       
             \xlmr              & N  & \ding{51}  & \ding{55} &  67.01 \footnotesize{$\pm$ 1.69}  & 69.01 \footnotesize{$\pm$ 0.55}	 &  67.51 \footnotesize{$\pm$ 1.59}  &  67.62 \footnotesize{$\pm$ 0.42}  &  68.0 \\
             \xlmr + Adapters   & N  & \ding{51}  & \ding{55} &  68.05 \footnotesize{$\pm$ 0.35}  & 68.02 \footnotesize{$\pm$ 1.11}	 &  66.99 \footnotesize{$\pm$ 1.01}  &  66.00 \footnotesize{$\pm$ 0.95}  &  67.0 \\
             \bottomrule
        \end{tabular}
    }
    \caption{Test R-Precision ($\mathrm{RP}$, \%) results $\pm$ std.\ deviation over 3 runs with different random seeds. E2E: End-to-End Fine-Tuning (FT). +Adapters: Updating only Adapter layers and classification head during FT. \#M: number of models fine-tuned. MT shows if machine-translated documents are used. BS+SL shows if teacher-student Boot-Strapping with Soft Labels is used.}
    \label{tab:appendix_fr_results}
\end{table*}

\section{Soft label quality}
\label{sec:main_analysis}

In Figure~\ref{fig:main_violins}, we estimate the quality of soft labels via the absolute differences between \emph{gold} and \emph{soft} labels (the latter predicted by the multilingual teacher model), separately per document subset (original in English, machine-translated in target languages, additional unlabelled documents in the target languages), for all the languages considered by the student. We compute differences as the averaged Mean Absolute Error (MAE) across documents and labels in each document subset:
\begin{equation}
    \overline{\mathrm{Diff}} = \frac{1}{N \times L} \sum_{n=1}^{N} \sum_{l=1}^{L} |G_{nl} - S_{nl}|
\end{equation}
\noindent where $N\!=\!12,000$ is the number of documents 
of the subset, $L\!=\!127$ is the number of labels, $G_{nl}\in\{0,1\}$ are the \emph{gold} labels for the $n$-th document, i.e., 1 when the $l$-th label is assigned to the $n$-th document and 0 otherwise, $S_{nl}\in[0,1]$ are the \emph{soft} labels (probabilities) for the  $n$-th document.
We observe that the quality of the soft labels vastly varies both across document subsets (considering the mean difference reported per violin with a thick blue horizontal line), and across documents within the subset (distribution in each violin).

The average differences ($\overline{\mathrm{Diff}}$) per language (source or target) fully correlate with the performance of the student model in the respective language, measured in RP, as reported in Table~\ref{tab:results}. Specifically, soft labels for French documents (machine-translated or unlabelled) are more accurate ($\overline{\mathrm{Diff}}\simeq0.25$) compared to the rest, i.e.,  $\overline{\mathrm{Diff}}\simeq0.45$ for German, and $\overline{\mathrm{Diff}}\simeq0.60$ for Greek. These results (soft label quality) seem to justify the performance improvement in French when the multilingual student is used, compared to the monolingual and bilingual students, and the performance deterioration in German and Greek. These results could also be affected by the quality of MT (Table~\ref{tab:nmt}). 

Based on these findings, we acknowledge that teacher-student bootstrapping should be reconsidered in future work with respect to the quality of translations and soft labels. Improvements could include discarding documents with very uncertain soft labels (probabilities), e.g., very close to a threshold (e.g., $t=0.5$), or weighing training documents by the certainty of their soft labels. Similarly, one could possibly filter out exceptionally low quality translations, e.g., detected via language modeling metrics (e.g., perplexity) using a pre-trained language model of the target language.\vspace{-1mm}

\section{Additional Results}
\label{sec:extra_results}

In this section, we show results of the same experiments described in Sections~\ref{sec:methods} and Section~\ref{sec:results}, but with different source languages (instead of English), excluding teacher-student models.  Overall, we can draw very similar conclusions as with Table~\ref{tab:results}, with a few exceptions, e.g., Translate-Test is worse for German and Greek when the source language is French, which is possibly related to the translation quality across these language pairs. 

In Table~\ref{tab:teacher_results}, we also present results for the Teacher models for all Student models presented in Table~\ref{tab:results}. We observe that in all (except one) cases (settings and languages), the Student model is comparable or outperforms the Teacher one.

\begin{table*}[ht!]
    \centering
    \resizebox{0.8\textwidth}{!}{
        \begin{tabular}{llcc|c|ccc|c}
        \hline
             & & & & \bf Source & \multicolumn{3}{c|}{\bf Target Languages} & \bf Target\\
             \bf Model & \bf \#M & \bf MT & \bf BS+SL  & \bf el & \bf de & \bf fr & \bf en & \bf Avg\\
             \hline
             \hline
             \multicolumn{9}{c}{\textit{\textbf{Zero-shot Cross-lingual FT (No labeled data in target languages)}}} \\
             \hline
             \hline
             \multicolumn{9}{l}{\textit{\textbf{Cross-lingual FT (Greek Only)}}} \\ 
             \hline
             \xlmr              & 1  & \ding{55} & \ding{55} &  64.57 \footnotesize{$\pm$ 0.39} & 46.30 \footnotesize{$\pm$ 3.23}  &  43.09 \footnotesize{$\pm$ 1.37}  &  41.54 \footnotesize{$\pm$ 2.02}  &  43.6 \\ 
             \xlmr + Adapters   & 1  & \ding{55} & \ding{55} &  64.86 \footnotesize{$\pm$ 0.19} & 49.89 \footnotesize{$\pm$ 3.81}  &  48.56 \footnotesize{$\pm$ 4.28}  &  47.98 \footnotesize{$\pm$ 4.75}  &  48.8 \\ 
             \hline
             \multicolumn{9}{l}{\textit{\textbf{Translate Test documents to Target language}}} \\
             \hline
             \xlmr              & 1  & \ding{51} & \ding{55} &  64.57 \footnotesize{$\pm$ 0.39} & 64.69 \footnotesize{$\pm$ 0.49}	&  64.59 \footnotesize{$\pm$ 1.53}  &  64.62 \footnotesize{$\pm$ 0.48}  &  64.6 \\ 
             \xlmr + Adapters   & 1  & \ding{51} & \ding{55} &  64.86 \footnotesize{$\pm$ 0.19} & 65.41 \footnotesize{$\pm$ 1.13}	&  62.89 \footnotesize{$\pm$ 0.95}  &  64.88 \footnotesize{$\pm$ 0.50}  &  64.2 \\ 
             \hline
             \multicolumn{9}{l}{\textit{\textbf{Translate Train documents to Target language}}} \\ 
             \hline       
             \xlmr              & N  & \ding{51}  & \ding{55} &  64.57 \footnotesize{$\pm$ 0.39} & 65.29 \footnotesize{$\pm$ 1.51}	 &  64.31 \footnotesize{$\pm$ 2.27}  &  64.77 \footnotesize{$\pm$ 1.30}  &  64.8 \\
             \xlmr + Adapters   & N  & \ding{51}  & \ding{55} &  64.86 \footnotesize{$\pm$ 0.19} & 66.22 \footnotesize{$\pm$ 0.22}	 &  64.76 \footnotesize{$\pm$ 1.24}  &  65.80 \footnotesize{$\pm$ 1.56}  &  65.6 \\
             \bottomrule
        \end{tabular}
    }
    \caption{Test R-Precision ($\mathrm{RP}$, \%) results $\pm$ std.\ deviation over 3 runs with different random seeds. E2E: End-to-End Fine-Tuning (FT). +Adapters: Updating only Adapter layers and classification head during FT. \#M: number of models fine-tuned. MT shows if machine-translated documents are used. BS+SL shows if teacher-student Boot-Strapping with Soft Labels is used.}
    \label{tab:appendix_el_results}
    \vspace{-5mm}
\end{table*}

\begin{table*}[ht!]
    \centering
    \resizebox{0.9\textwidth}{!}{
        \begin{tabular}{llcc|c|ccc|c}
            \hline
            & & &  & \bf Source & \multicolumn{3}{c|}{\bf Target Languages} & \bf Target\\
             \bf Model & \bf \#M & \bf MT & \bf BS+SL  & \bf en & \bf de & \bf fr & \bf el & \bf Avg\\
             \hline
             \multicolumn{9}{l}{\textit{\textbf{Bilingual Teacher-Student (jointly FT on Source documents and their translations in a Target language)}}} \\ 
             \hline 
             \xlmr (E2E) (Teacher)        & 4  & \ding{51} & \ding{55} & 68.2 \footnotesize{$\pm$ 0.8}               &  67.1 \footnotesize{$\pm$ 0.71}  &  65.8 \footnotesize{$\pm$ 0.4}  &  65.0 \footnotesize{$\pm$ 1.5}  &  66.0 \\
             \xlmr (E2E) (Student)  & 4  & \ding{51} & \ding{51} & \textbf{69.1} \footnotesize{$\pm$ 1.3}  & \textbf{67.4} \footnotesize{$\pm$ 0.1}  &  66.1 \footnotesize{$\pm$ 0.3}  &  65.0 \footnotesize{$\pm$ 0.4}  &  66.1 \\
             \hline
             \xlmr +Adapters (Teacher)     & 4  & \ding{51} & \ding{55} & 68.8 \footnotesize{$\pm$ 0.1}               &  66.7 \footnotesize{$\pm$ 0.7}  &  66.9 \footnotesize{$\pm$ 1.2}  &  65.4 \footnotesize{$\pm$ 1.1}  &  66.3 \\
             \xlmr +Adapters (Student) & 4  & \ding{51} & \ding{51} & 67.8 \footnotesize{$\pm$ 1.3}  &  66.9 \footnotesize{$\pm$ 0.3}  &  67.6 \footnotesize{$\pm$ 1.2}  &  \textbf{67.9} \footnotesize{$\pm$ 0.1}  &  \bf 67.5 \\
             \hline
             \multicolumn{9}{l}{\textit{\textbf{Multilingual Teacher-Student (jointly FT on Source documents and their translations in all Target languages)}}} \\ 
             \hline
             \xlmr (E2E) (Teacher)      & 1  & \ding{51} & \ding{55} & 59.2 \footnotesize{$\pm$ 2.5}  &  57.0 \footnotesize{$\pm$ 2.6}  &  67.6 \footnotesize{$\pm$ 1.4}  &  46.0 \footnotesize{$\pm$ 2.0}  &  56.9 \\
             \xlmr (E2E) (Student)   & 1  & \ding{51} & \ding{51} & 62.3 \footnotesize{$\pm$ 1.6}  &  60.9 \footnotesize{$\pm$ 0.3}  &  66.8 \footnotesize{$\pm$ 0.2}  &  48.4 \footnotesize{$\pm$ 0.3}  &  58.7 \\
             \hline
             \xlmr +Adapters (Teacher)   & 1  & \ding{51}  & \ding{55} & 64.3 \footnotesize{$\pm$ 1.7}  &  60.0 \footnotesize{$\pm$ 0.2}  &  49.3 \footnotesize{$\pm$ 2.3}  &  49.3 \footnotesize{$\pm$ 2.3}  &  52.9 \\
             \xlmr +Adapters (Student) & 1  & \ding{51}  & \ding{51} & 65.0 \footnotesize{$\pm$ 0.2}  &  62.6 \footnotesize{$\pm$ 0.2}  &  \textbf{68.7} \footnotesize{$\pm$ 0.8}  &  50.5 \footnotesize{$\pm$ 0.0}  &  60.6 \\
             \bottomrule
        \end{tabular}
    }
    \caption{Test R-Precision ($\mathrm{RP}$, \%) results $\pm$ std.\ deviation over 3 runs with different random seeds. E2E: End-to-End Fine-Tuning (FT). +Adapters: Updating only Adapter layers and classification head during FT. \#M: number of models fine-tuned. MT: machine-translated documents used. BS+SL: Boot-Strapping with Soft Labels.
    }
    \label{tab:teacher_results}
    \vspace{-7mm}
\end{table*}

\end{document}